\documentclass[onefignum,onetabnum]{siamonline220329}

\usepackage{lipsum}
\usepackage{amsfonts}
\usepackage{bm}
\usepackage{graphicx}
\usepackage{epstopdf}
\usepackage[export]{adjustbox}
\usepackage[caption=false]{subfig} %
\usepackage{algorithmic}
\usepackage{pgfplots}
\usepackage{layouts}
\usepackage{mathtools}
\usepackage{etoolbox}

\ifpdf
  \DeclareGraphicsExtensions{.eps,.pdf,.png,.jpg}
\else
  \DeclareGraphicsExtensions{.eps,.pdf,.png,.jpg}
\fi

\DeclarePairedDelimiterX\set[2]{\{}{\}}
  {#1 \mathrel{}\mathclose{}\delimsize|\mathopen{}\mathrel{} #2}

\usepackage{enumitem}
\setlist[enumerate]{leftmargin=.5in}
\setlist[itemize]{leftmargin=.5in}

\newsiamremark{remark}{Remark}
\newsiamremark{hypothesis}{Hypothesis}
\crefname{hypothesis}{Hypothesis}{Hypotheses}
\newsiamthm{claim}{Claim}

\headers{UQ in Machine Learning-Based Biomed Segmentation}{F. Terhag, P. Knechtges, A. Basermann, R. Tempone}

\title{Uncertainty Quantification in Machine Learning Based Segmentation:\\
A Post-Hoc Approach for Left Ventricle Volume Estimation in MRI}

\author{Felix Terhag{\thanks{Department for High-Performance Computing, German Aerospace Center (DLR), Institute of Software Technology, Cologne, Germany
  (\email{felix.terhag@dlr.de}, \email{philipp.knechtges@dlr.de}, \email{achim.basermann@dlr.de}).}} \thanks{Chair of Mathematics for Uncertainty Quantification, RWTH Aachen University, Aachen, Germany (\email{tempone@uq.rwth-aachen.de})} \and Philipp Knechtges\footnotemark[1] \and Achim Basermann\footnotemark[1]
\and Raúl Tempone\footnotemark[2] \thanks{Computer, Electrical and Mathematical Sciences and Engineering Division, KAUST, Thuwal, Saudi Arabia}}

\usepackage{amsopn}

\usepackage{placeins}
\usepackage{todonotes}

\ifpdf
\hypersetup{
  pdftitle={Uncertainty Quantification in Machine Learning Based Segmentation: A Post-Hoc Approach for Left Ventricle Volume Estimation in MRI},
  pdfauthor={F. Terhag, P. Knechtges, A. Basermann, R. Tempone}
}
\fi

\begin{document}

\maketitle

\begin{abstract}
Recent studies have confirmed cardiovascular diseases remain responsible for highest the death toll amongst non-communicable diseases. The accurate left ventricular (LV) volume estimation is critical for valid diagnosis and management of various cardiovascular conditions, but poses a significant challenge due to inherent uncertainties associated with the segmentation algorithms in magnetic resonance imaging (MRI).  Recent machine learning advancements, particularly U-Net-like convolutional networks, have facilitated automated segmentation for medical images, but struggles under certain pathologies and/or different scanner vendors and imaging protocols. This study proposes a novel methodology for post-hoc uncertainty estimation in the LV volume prediction using It\^{o} stochastic differential equations (SDEs) to model path-wise behavior for the prediction error. The model describes the area of the left ventricle along the heart's long axis. The method is agnostic to the underlying segmentation algorithm, facilitating its use with various existing and future segmentation technologies. The proposed approach provides a mechanism for quantifying uncertainty, enabling medical professionals to intervene for unreliable predictions. This is of utmost importance in critical applications such as medical diagnosis, where prediction accuracy and reliability can directly impact patient outcomes. The method is also robust to dataset changes, enabling application for medical centers with limited access to labeled data. Our findings highlight the proposed uncertainty estimation methodology's potential to enhance automated segmentation robustness and generalizability, paving the way for more reliable and accurate LV volume estimation in clinical settings as well as opening new avenues for uncertainty quantification in biomedical image segmentation, providing promising directions for future research.
\end{abstract}

\begin{keywords}
Machine learning, Uncertainty quantification, Cardiovascular MRI, It\^{o} stochastic differential equations, U-Net, Neural networks, Left ventricle volume estimation, Biomedical image segmentation, Convolutional neural networks

\end{keywords}

\begin{AMS}
68T07, 62P10, 92C55, 68T05, 65C20, 62M45
\end{AMS}

\section{Introduction}
Recent studies have confirmed cardiovascular diseases remain responsible for the highest death toll amongst non-communicable diseases~\cite{roth_global_diseases_2018}.
The left ventricle (LV) volume is an essential and well-studied parameter to help measure or asses many cardiovascular related conditions, including hypertension~\cite{missouris_echocardiography_1996}, cardiovascular mortality~\cite{verdecchia_prognostic_1998}, and heart failure~\cite{missouris_echocardiography_1996}. The gold standard for LV volume measurement is manually segmenting the LV by experienced investigators in short-axis view magnetic resonance images (MRIs)~\cite{perdrix_calc_LV_2011}, which is both a tedious and costly procedure. Short-axis MRIs consists of several slices along the short-axis of the heart.\\
Recent machine learning developments, notably U-Net-like convolutional networks~\cite{Long_FCN_2017}, have enabled better automatic segmentation for medical images; achieving almost expert-level segmentation for short-axis cardio MRIs in several public challenges~\cite{Acdc_2018, MnM_2021, isensee_nnu-net_2021}. However, the prediction quality tends to fall off for some pathologies~\cite{Acdc_2018} and has relatively poor generalizability across the various MRI scanner vendors and/or imaging protocols~\cite{MnM_2021}. Accurate and automated machine learning segmentation provide beneficial results, but particularly for critical applications, i.e., medical diagnosis, accurate uncertainty descriptors are essential to assist medical experts to infer reasonable outcomes from potentially unreliable predictions.\\
Therefore, this study  proposes a post-hoc uncertainty estimation method for LV volume prediction that is agnostic to the underlying segmentation algorithm. Our method takes inspiration from~\cite{caballero_quantifying_2021}, where a similar approach with a different base model was chosen to quantify the uncertainty in wind power forecasting. The proposed approach can be used with existing and future segmentation algorithms, and requires only eight parameters to be fit on datasets. Thus, the proposed method can be applied for smaller datasets rather than retraining a neural network with several million parameters. \\
Automatic LV segmentation is a widely studied field, not least due to the segmentation challenges~\cite{Acdc_2018} and~\cite{MnM_2021}. Several segmentation algorithms have been developed, most employing U-Net neural network architecture~\cite{isensee_nnu-net_2021, Zhang2021Segmentation},  e.g. the ten best performing methods from the 2021 M\&Ms Challenge all relied on this architecture~\cite{MnM_2021}. Although several attempts have been proposed to improve the neural network prediction uncertainty, they usually involve changes to the architecture and/or training routine~\cite{kohl2018_prob_unet,gal2016dropout}. Thus, one cannot generally apply these methods onto current state-of-the-art methods and they would usually require at least as many training examples as retraining a neural network from scratch~\cite{kohl2018_prob_unet, gal2016dropout, blundell_bayes_by_backprop_2015}. Retraining neural networks is problematic for real-word applications since medical centers have different MRI machines and/or post-processing techniques, making it difficult to generalize over research centers~\cite{MnM_2021}. Thus, these techniques are not practical for smaller medical centers, which cannot obtain a sufficient number of labeled MRIs. The proposed approach to provide uncertainty quantification on a derived parameter of medical importance from segmented MRIs has, to our knowledge, not been previously reported.\\
The remainder of this paper is organized as follows; Section~\ref{sec:application} introduces the left ventricle volume prediction problem, and we subsequently describe the proposed modelling approach in Section~\ref{sec:Modelling}. Sections ~\ref{ssec:model_inner},~\ref{ssec:wall_modelling}, and \ref{ssec:combining_sde_j} differentiate between modelling inner and outer slices and how to combine these, respectively. Section~\ref{sec:likelihood} describes fitting model parameters using maximum likelihood, with experimental results presented in Section~\ref{sec:exp_results}. Finally, Section~\ref{sec:conclusion} summarizes and concludes the paper.%
\begin{figure}[h!]
\label{fig:schematic_proc}
\centering
\subfloat{\tikz[remember
picture]{\node(1AL){\includegraphics[width=3cm,margin=0cm 0.55cm 0cm 0cm]{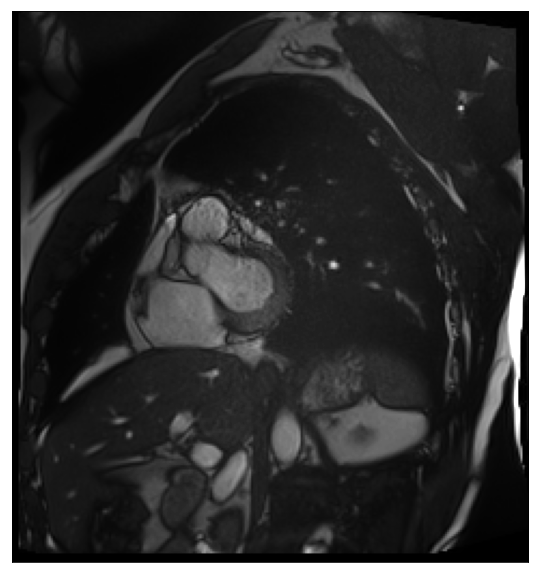}};}}%
\hspace*{1cm} 
\subfloat{\tikz[remember picture]{\node(1AM){\includegraphics[width=3cm,margin=0cm 0.55cm 0cm 0cm]{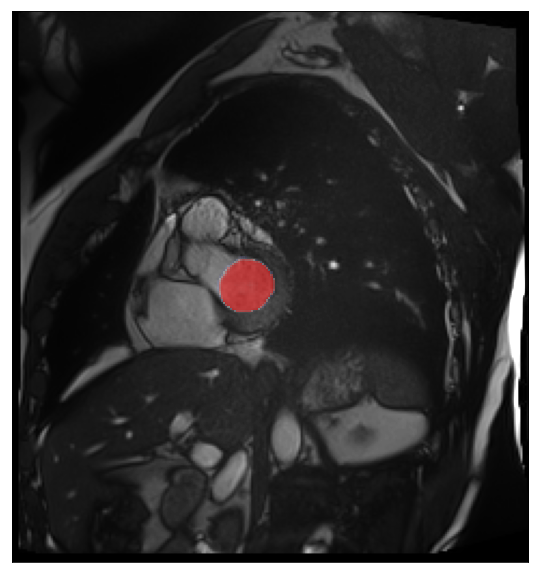}};}}
\hspace*{1cm}%
\subfloat{\tikz[remember picture]{\node(1AR){\input{figures/distributional_prediction.pgf}};}}
\caption{Proposed approach to model left ventricle (LV) volume from short-axis cardiac MRI (the example image is from the ACDC Dataset~\cite{Acdc_2018}). (1) An arbitrary algorithm, segments the left ventricle. (2) Our method creates a distributional volume prediction from deterministic segmentations.}
\tikz[overlay,remember picture]{\draw[-latex,thick] ([shift={(0,0.26)}]1AL.east) -- ([shift={(0,0.26)}]1AL-|1AM.west)
node[midway,below,text width=0.5cm]{(1)};} 
\tikz[overlay, remember picture]{\draw[-latex,thick] ([shift={(0,0.26)}]1AM.east) -- ([shift={(0,0.26)}]1AM-|1AR.west)
node[midway,below,text width=0.5cm]{(2)};} 
\tikz[overlay, remember picture]{\draw[black,thick] ([shift={(0,0.1)}]1AM.north west) -- ([shift={(8.1,0.1)}]1AM.north west)-- ([shift={(8.1,0)}]1AM.south west)-- (1AM.south west) -- ([shift={(0,0.1)}]1AM.north west)}
\end{figure}%

\section{Volume Prediction from Cardio MRIs}
\label{sec:application}Short-axis cardiac MRIs comprise several parallel slices recorded perpendicular to the long-axis, i.e., the plane intersecting the base of the heart and the apex. In particular, these slices show the LV cross-sectional area. Typical MRI spatial resolution is much higher than resolution along the slices, e.g. spatial resolution $\approx 0.85 - 1.32$~mm, whereas slice thickness can be as much as ten times higher. We are interested in LV volumes for end-diastolic and end-systolic phases to calculate essential parameters for cardiac diagnosis~\cite{MnM_2021}.\\
This study employs the ACDC~\cite{Acdc_2018} and M\&Ms datasets~\cite{MnM_2021}, containing short-axis view cardiac MRIs, with manually annotated end-systolic and end-diastolic MRI for 100 and 150 patients, respectively. Both datasets also include patients with numerous pathologies. Patients in the ACDC datasets are evenly distributed over five characteristics and pathologies classes, e.g., patients with normal cardiac anatomy and function, patients with systolic heart failure and infarction, and patients with abnormal right ventricle. In contrast, the M\&Ms dataset includes seven classes for the most frequent pathologies, with a separate class for healthy patients and one for other pathologies which are not represented in the most frequent pathologies. The M\&Ms dataset contains MRIs from different medical centers with different MRI machine vendors and imaging protocols, whereas the ACDC dataset includes MRIs from a single medical center with the same imaging protocol.  Thus the M\&Ms dataset provides not only higher pathologic diversity, but also MRI vendor and imaging protocol diversity.\footnote{More detailed descriptions for the ACDC and M\&M datasets can be found in~\cite{Acdc_2018,MnM_2021}.}\\
Current state-of-the-art automatic segmentation methods for MRIs are convolutional neural networks with U-Net architecture~\cite{Acdc_2018, MnM_2021}. The U-Net architecture was introduced by Long et al. in~\cite{Long_FCN_2017}. Following the characteristic U-Net approach, the first half condenses information using convolutional and subsequently by pooling layers, reducing spatial information while typically increasing the number of filters. The second half comprises upsampling deconvolutions, such that the last level produces images with the same resolution as the input image. Downsampling and upsampling paths are connected with skip connections, where feature maps from the downsampling path are passed to the corresponding step in the upsampling path. Isensee et al. ~\cite{isensee_nnu-net_2021} proposed the nnU-net framework for automatic architecture search specifically for medical segmentation tasks, and this method won both ACDC and M\&Ms segmentation challenges. Therefore, we employ a pre-trained nnU-Net version, trained on the M\&Ms dataset since this is the richer dataset, and use the ACDC dataset for evaluation, hence we test the method on many unseen examples. This also enables realistic applications, where practitioners were typically limited to data from their research center and use a pretrained net that was trained on a very diverse dataset. The nnU-Net architecture~\cite{isensee_nnu-net_2021} commonly employs an ensemble of five neural networks for inference. For simplicity, we take a single net since this only requires approximately one fifth of the compute time. The pre-trained net can predict four classes: \emph{left ventricle}, \emph{right ventricle}, \emph{myocard}, and \emph{background}, i.e., the prediction for every voxel $i,j$ contains four values. We focused on the LV volume since this is the most important medical parameter~\cite{missouris_echocardiography_1996}, but the method could also be applied to the other classes. \\
Hence, the only information we get from the outputs of the neural net is one value $o_{i,j}$ per voxel. To build a model directly from the output of the neural network, one could consider each output $o_{i,j}$ as the predicted probability that voxel $(i,j)$ belongs to the LV. To mimic this, the volume can be modeled as a sum of independent Bernoulli random variables $O_{i,j}$ with probability $o_{i,j}$ describing whether or not voxel $(i,j)$ belongs to the LV. The predicted volume for one MRI slice $\textit{Vol}_{output}$ can be described as
\begin{equation} \label{eq:model_RV_naive}
\textit{Vol}_{output}=\nu \cdot \sum_i \sum_j O_{i,j},
\end{equation}
where $\nu$ is the volume of one voxel. The expected value is the sum of the probabilities multiplied by the the volume of one voxel: $\mathbb{E}[\textit{Vol}_{output}]=\nu \sum_i \sum_j o_{i,j}$. This approach reveals the difficulties of the neural network to accurately predict the uncertainties. Too see this, we compare the standard deviation from \eqref{eq:model_RV_naive} with observed error from ground-truth labels, as shown in Figure~\ref{fig:error_vs_std}. To obtain the ground-thruth labels, experts manually assigned each voxel to the corresponding class. 
\begin{figure}[tbhp]
    \begin{center}
        \input{figures/error_vs_std.pgf}
    \end{center}
    \label{fig:error_vs_std}
    \caption{Left ventricle area absolute error and predicted standard deviation (from \eqref{eq:model_RV_naive}); outer and inner slices are shown as orange and blue, respectively}
\end{figure}
Not only is the observed error several magnitudes larger than the modeled variation, but there is no apparent correlation between higher modeled variation and higher absolute error. Of particular concern, is that some examples exhibit very low predicted standard deviation but high observed error. For example, the highest absolute error slice has one of the smallest modeled standard deviations. Thus, the outputs of the neural network cannot accurately capture the underlying uncertainties, consistent with previous studies showing that deep neural networks for classification~\cite{guo_calibration_2017} and segmentation problems~\cite{Rousseau_Calibration_Segmentation_2021} are overly confident. Therefore, a different modeling approach is required.

\section{Modelling the Volume Under Uncertainty}\label{sec:Modelling} Section~\ref{sec:application} establishes that treating neural network outputs as predicted probabilities results in a poor model. Therefore, we only use the expected value of \eqref{eq:model_RV_naive} as a point estimate for the volume prediction for each slice. Thus, if MRI $j\in\{1,\ldots, M\}$ contains $N_j+1$ slices, we obtain $N_j+1$ deterministic point predictions $p_{0}^{(j)},\ldots ,p_{N_j}^{(j)}$, with $M$ being the number of hearts in the dataset.\\
Even human experts find slices through the apical or basal heart sections, which we call outer slices, to be the hardest to segment. In those slices, it is often unclear whether a voxel belongs to the LV or adjacent tissue, and this is echoed in neural network predictions. Although outer slices only comprise $25\%$ of all slices, they account for 27 of 31 slices with absolute error $> 600$~mm$^2$.  We define the \emph{outer slices} as two MRI slices where neighboring slices comprise one zero and one nonzero prediction. Since none of the predictions are effectively zero, we define zero predictions where the area is smaller than a threshold $\epsilon$. Figure~\ref{fig:zero_epsilon} shows the deterministic predictions for each slice. We see, that $\epsilon = 10$~mm$^2$ is a reasonable threshold to separate very low predictions from the majority of predictions. 
\begin{figure}[tbhp]
    \begin{center}
        \input{figures/histogram_all_preds_zero_epsilon.pgf}
    \end{center}
    \label{fig:zero_epsilon}
    \caption{Deterministic predictions of the left ventricle area for all slices in the dataset. The chosen threshold for zero predictions is shown in red. This is necessary as the neural network never predicts exactly zero.}
\end{figure}
We then preprocess the data such that all predictions start and end with a zero prediction and the next prediction is non-zero, which requires appending zeros or truncating consecutive zero predictions. Truncating consecutive zero predictions does not lose information, since the corresponding labeled data also contains no LV volume, hence $p_{0}^{(j)}=p_{N_j}^{(j)}=0$, where $p_{1}^{(j)},\ldots,p_{N_j-1}^{(j)}>0$ for all MRIs  $j\in\{1,\ldots, M\}$.
\subsection{Modelling the inner slices}\label{ssec:model_inner} 
Section \ref{sec:application} establishes that the error behaves fundamentally differently for outer and inner heart slices. Although inner slices generally exhibit smaller deviation from the prediction, outer slices exhibit more categorical error, such that the full predicted area can be assigned to the wrong class. Therefore, we model the inner and outer sections independently. Excluding the first and last slices, we model the inner slices as follows:
Neighboring slices for any given heart have an inherent dependency due to locality, which we want to reflect by constructing an SDE for the slice area $X_t$ over a fake time $t$, with $t$ ranging over the depth of the heart and $X_t$ being measured at integer values of $t$. More specifically, we model the area $X_t$ over the depth of the heart $t$ by the stochastic process $X=\{X_t, t\in [0,T]\}$,
\begin{equation}\label{eq:SDE_general}
    \begin{cases}
        dX_t = a(X_t, p_t, \dot{p}_t, \bm{\theta})dt+b(X_t, p_t, \dot{p}_t, \bm{\theta})dW_t,\\
        t\in [0,T], X_0 > 0,
    \end{cases}
\end{equation}
where  $a(\cdot, p_t, \dot{p}_t,\bm{\theta}): \mathbb{R}^{+}\rightarrow \mathbb{R}$ is the drift function; $b(\cdot, p_t, \dot{p}_t,\bm{\theta}):\mathbb{R}^{+}\rightarrow \mathbb{R}^{+}$ is the diffusion function; $\bm{\theta}$ is a parameter vector; $(p_t)_{t\in [0,T]}$ a depth-dependent $\mathbb{R}^{+}$-valued deterministic function with time derivative $(\dot{p}_t)_{t\in[0,T]}$; and $\{W_t, t\in [0,T]\}$ is a standard real-valued Wiener process.
Thus, $(p_t)_{t\in [0,T]}$ is the linear interpolation of the deterministic point predictions omitting superscript $(j)$. Following~\cite{caballero_quantifying_2021}, we choose the drift function,
\begin{equation}\label{eq:drift}
    a(X_t, p_t, \dot{p}_t, \bm{\theta})=\dot{p}_t - \theta_t(X_t-p_t),
\end{equation}
to make the process $X_t$ bias-free with respect to the predictions $p_i$. More precisely, the process will satisfy $\mathbb{E}[X_t]=p_t$, if $\mathbb{E}[X_0]=p_0$ holds initially. We also want the process to be mean-reverting at least with rate $\theta_0>0$, hence $\theta_t>\theta_0$.\\
The LV area over the depth of the heart should not become negative, i.e., the state space for $X_t$ is $\mathbb{R}^{+}$. Let $\bm{\theta}=(\theta_0, \alpha)$ and $\alpha>0$ be parameters controlling the path variability. Then, the state-dependent diffusion 
\begin{equation}
    b(X_t, \bm{\theta}) = \sqrt{2\alpha\theta_0 X_t},
\end{equation}
in combination with
\begin{equation*}
    \theta_t\geq \frac{\alpha \theta_0-\dot{p}_t}{p_t},
\end{equation*}
ensures that $X_t \in \mathbb{R}^+$ almost surely. Hence we choose
\begin{equation}\label{eq:theta_t}
\theta_t=\max \left(\theta_0, \frac{\alpha\theta_0-\dot{p}_t}{p_t}\right).
\end{equation}
To give proof of the last claim, we will prove the following theorem.
\begin{theorem}\label{theorem1}
Assume a probability space rich enough to accommodate Brownian motion $W_t$ with $t\in[0,T]$ and a thereof independent random variable $X_0$ with $X_0 > 0$ a.s.; function $p: [0,T] \to \mathbb{R}^{>0}$ with piecewise continuous derivative; and $\alpha, \theta_0 > 0$ and a piecewise continuous function $\theta: [0,T]\to \mathbb{R}^{>0}$, with
\begin{equation}\label{eqn:proofboundontheta}
    \theta_t \geq \max \left(0, \frac{\alpha\theta_0-\dot{p}_t}{p_t}\right)\, .
\end{equation}
Then, the stochastic differential equation
\begin{equation}\label{eq:SDE_X}
    dX_t = (\dot{p}_t-\theta_t(X_t-p_t))dt+\sqrt{2\alpha \theta_0 X_t}\,dW_t
\end{equation}
has a unique strong solution, and $0 < X_t < \infty$ holds for all $t\in [0,T]$ almost surely.
\end{theorem}
\begin{proof}
Without loss of generality, assume $\sqrt{2\alpha \theta_0 X_t}$ to be padded by zeros for negative values of $X_t$.  Then by potentially limiting us to the segments of $[0,T]$ where $\dot{p}_t$ and $\theta_t$ are continuous, applying~\cite[p.~59]{Skorokhod2017} successively proves the existence of a strong solution with $X_t < \infty$ for each of these segments, and the Yamada--Watanabe theorem~(\cite[p.~291]{Karatzas1998},\cite{Yamada1971}) confirms the uniqueness of this solution. Thus, we only need to show that $X_t > 0$ almost surely. We follow the McKean argument as given in~\cite[p.~23]{Alfonsi2015} and~\cite[Lemma~4.2]{Mayerhofer2011}. 
At first, we define a stopping time $\tau_0(\omega) = \inf \set{t}{X_t(\omega) = 0}$ with $\tau_0(\omega) = \infty$ for $\set{t}{X_t(\omega) = 0} = \emptyset$. Then, using Itô's formula for $f(x) = \log(x)$ it follows
\begin{align*}
    X_t =& X_0 \exp\left(\int_0^t \frac{\dot{p}_s + \theta_s p_s - \alpha \theta_0}{X_s}\, ds
        - \int_0^t \theta_s\, ds + \int_0^t \sqrt{\frac{2\alpha\theta_0}{X_s}}\, dW_s\right)\, ,
\end{align*}
for $t\in[0,\tau_0)$, and using~\eqref{eqn:proofboundontheta} yields
\begin{align*}
    X_t \geq& X_0 \exp\left(
        - \int_0^t \theta_s\, ds + \int_0^t \sqrt{\frac{2\alpha\theta_0}{X_s}}\, dW_s\right)\, .
\end{align*}
Thus, the set $\set{\omega}{\tau_0(\omega) \neq \infty}$ is a subset of
\begin{align*}
    A\coloneqq& \set*{\omega}{\lim_{t\to T\wedge \tau_0(\omega)} \int_0^{t} \sqrt{\frac{2\alpha\theta_0}{X_s(\omega)}}\, dW_s(\omega) =  -\infty}\, .
\end{align*}
However, $M_{t} \coloneqq \int_0^{t} \sqrt{\frac{2\alpha\theta_0}{X_s}}\, dW_s$ is a continuous local martingale on $[0,\tau_0)$, for which  ~\cite[Lemma~4.2]{Mayerhofer2011} yields that $A$ has probability zero. Hence, $\tau_0 = \infty$ almost surely.
\end{proof}
With a simple change of variables, we obtain a model for the prediction error $V_t = X_t-p_t$ given by
\begin{equation} \label{eq:SDE_V}
    \begin{cases}
        dV_t = -\theta_t V_t dt+\sqrt{2\alpha \theta_0 (V_t+p_t)}dW_t\\
        t\in [0,T], V_0 = v_0\geq -p_0.
    \end{cases}
\end{equation}
This formulation provides us a compact description for the prediction error and Theorem~\ref{theorem1} confirms that with choosing appropriate  $\theta_t$ we obtain a unique strong solution for the SDE which does not leave $\mathbb{R}^{+}$ almost surely. Hence the modeled LV area over the inner slices does not become zero almost surely.

\subsection{Modelling the outer slices}\label{ssec:wall_modelling}
As discussed above, outer slices are especially hard to predict (see Fig.~\ref{fig:error_vs_std}). Particularly large errors occur due to systematic error in the outer slices, and hence some outer slices are falsely segmented as LV, or the other way around, some outer slices show LV where the nnU-Net assumes only LV adjacent tissue is visible. The SDE is not designed to incorporate large jumps, particularly where the prediction is near zero. The mean-reverting parameter $\theta_t$ is large when $p_t$ is close to zero, as shown in \eqref{eq:theta_t}. This causes the trajectories to stay close to prediction $p_t$. Therefore, outer MRI slices must be modeled separately due to this systematic error and the SDE's inability to handle those errors well. Critical slices are $p_{0}=0$ and $p_{1}>0$ and $p_{N}=0$ and $p_{N-1}>0$. We treat the first and last two slices equivalently with the same parameters, and hence it suffices to only consider the first two slices. We want the model to be able to cover three mutually exclusive cases as follows:
\vspace{1mm}\begin{enumerate}
     \item $x_0=0$, $x_1>0$\qquad (no jump)
     \item $x_0>0$, $x_1>0$\qquad (jump up)
     \item $x_0=0$, $x_1=0$\qquad (jump down).
\end{enumerate}
{\bf Case 1}. The first case exhibits relatively small error. Slice zero is modeled with no volume and slice one contains LV volume. In this case the model agrees with the neural network. The subsequent cases describe systematic errors that could occur in the outer slices: {\bf Case 2}. We model the possibility that, contrary to the neural network prediction, there is already LV volume in slice zero, which often occurs when the neural network falsely classifies the LV in slice zero as surrounding tissue. {\bf Case 3}. We model the possibility that there is no LV volume in slice one. Case 2 \& 3 errors can have a large effect on the final prediction. To be exhaustive in the logical combinations, the fourth possible case ($x_0>0$ and $x_1=0$) is not modeled since we assume the LV to be a connected volume, which negates the possibility for zero predictions between non-zero predictions.\\
Let $X_0$ and $X_1$ be random variables describing the volume at slice 0 and 1, respectively. We choose a hierarchical approach based on a categorical distribution to model their behavior. This distribution models which of the three depicted cases occurs: jump-up, jump-down, or no-jump with probability $\lambda_u$, $\lambda_d$, and $1-\lambda_u-\lambda_d$, respectively; where $\lambda_d,\lambda_u\in[0,1]$ and $\lambda_u+\lambda_d\leq1$. After determining the case in the first layer of the hierarchical model, we draw from a specific distribution for this case, which then composes the second layer of the hierarchical model. We cannot model $X_0$ and $X_1$ independently because we have to prevent $x_0$ being nonzero while $x_1=0$, which would lead to the model having a zero volume in slice one between two nonzero slices. The distribution also must depend on the predicted volume $p_0+p_1=p_1$ in the first two slices since we want the model to inherit model properties for the inner slices, more specific, that the expected volume matches the segmentation algorithm prediction. These considerations lead to a probability density function
\begin{equation}\label{eq:pdf_jump}
    f_{X_0,X_1}(x_0,x_1\mid \Theta_J,p_1)=
    \begin{cases} 
        \lambda_d & \text{for } x_0=x_1=0\\
        \lambda_u \cdot f_{u0}(x_0\mid\beta_{u0},p_1)\cdot f_{u1}(x_1\mid\beta_{u1},p_1) & \text{for } x_0,x_1>0\\
        (1-\lambda_d-\lambda_u)\cdot f_n(x_1\mid\beta_n,p_1) & \text{for } x_0=0, x_1>0,
    \end{cases}
\end{equation}
where $f_{u0}$, $f_{u1}$, and $f_n$ are density functions for the jump-up and no-jump case distributions, respectively, defined by predicted volume $p_1$ and some parameter $\beta$. There is no distribution for the jump-down case since the result $x_0=x_1=0$ is deterministic. For simplicity, we abbreviate the notation $\Theta_J =  \left(\beta_n,\allowbreak \beta_{u0},\allowbreak \beta_{u1},\allowbreak \lambda_u,\allowbreak \lambda_d\right)$ as all parameters necessary for the jump distribution. In order to keep the cases in \eqref{eq:pdf_jump} disjunct, we require $f_{u0}(0\mid\beta_{u0})  = f_{u1}(0\mid\beta_{u1}) =f_n(0\mid\beta_n)=0$.\\
The probability density function \eqref{eq:pdf_jump} can also be expressed as
\begin{equation}\label{eq:pdf_jump_closed}
\begin{split}
    f_{X_0,X_1}(x_0,x_1\mid \Theta_J) =& \lambda_d\,\delta(x_0)\delta(x_1)+ \lambda_u \cdot f_{u0}(x_0\mid\beta_{u0},p_1)\cdot f_{u1}(x_1\mid\beta_{u1},p_1)\\
    &+(1-\lambda_d-\lambda_u)\cdot f_n(x_1\mid\beta_n,p_1)\delta(x_0)
\end{split}
\end{equation} 
with the Dirac delta distribution $\delta$. Hence, we can derive the marginal distributions 
\begin{equation}\label{eq:pdf_x0}
\begin{split}
    f_{X_0}\left(x_0\mid \Theta_J\right) =& \int_0^{\infty} f_{X_0,X_1}(x_0,x_1\mid \Theta_J) dx_1\\
    =&\lambda_d\cdot \delta(x_0)+(1-\lambda_u - \lambda_d)\cdot \delta(x_0) + \lambda_u\cdot  f_{u0}(x_0\mid\beta_{u0},p_1)\\
    =&(1-\lambda_u)\cdot  \delta(x_0) + \lambda_u\cdot  f_{u0}(x_0\mid\beta_{u0},p_1)
\end{split}
\end{equation}
and 
\begin{equation}\label{eq:pdf_x1}
\begin{split}
    f_{X_1}\left(x_1\mid \Theta_J\right) =& \int_0^{\infty} f_{X_0,X_1}(x_0,x_1\mid \Theta_J) dx_0\\
    =&\lambda_d\cdot \delta(x_1)+(1-\lambda_u - \lambda_d) \cdot f_{n}(x_1\mid\beta_{n},p_1) + \lambda_u \cdot f_{u1}(x_1\mid\beta_{u1},p_1);
\end{split}
\end{equation}
and their expected values 
\begin{equation}\label{eq:ev_x0}
        \mathbb{E}[X_0] = \lambda_u \cdot \int_0^{\infty}x_0 \cdot f_{u0}(x_0\mid\beta_{u0},p_1) dx0
\end{equation}
and 
\begin{equation}\label{eq:ev_x1}
\begin{split}
    \mathbb{E}[X_1] =& \lambda_u \cdot \int_0^{\infty}x_1\cdot  f_{u1}(x_1\mid\beta_{u1},p_1)dx_1\\
    &+(1-\lambda_u - \lambda_d) \cdot \int_0^{\infty}x_1\cdot f_{n}(x_1\mid\beta_{n},p_1)dx_1.
\end{split}
\end{equation}
To keep the model bias-free with respect to the prediction, expected values for the first two slices should be equal to the prediction in those slices, i.e., we require $\mathbb{E}[X_0+X_1]=p_0+p_1=p_1$. 
The model should have the same mean as the prediction $p_1$ in slice one if the simulation is nonzero, hence we require $\mathbb{E}_{X_1\sim f_{u1}}[X_1]=\mathbb{E}_{X_1\sim f_{n}}[X_1]=p_1$. With \eqref{eq:ev_x0}, \eqref{eq:ev_x1} and $\mathbb{E}\left[X_0\right]+\mathbb{E}[X_1]=\lambda_u \mathbb{E}_{X_0\sim f_{u0}}[X_0]+(1-\lambda_d) \cdot p_1$, we require
\begin{equation}\label{eq:calc_EX0_EX1}
    \lambda_u \mathbb{E}_{X_0\sim f_{u0}}[X_0]+(1-\lambda_d) \cdot p_1 =p_1
\end{equation}
and
\begin{equation}\label{eq:Eu0}
    \mathbb{E}_{X_0\sim f_{u0}}[X_0] =\frac{\lambda_d}{\lambda_u} p_1.
\end{equation}
Therefore, the desired mean $\mu_{u0}(p_1)$, $\mu_{u1}(p_1)$ and $\mu_n(p_1)$ for the distributions only depends on the prediction $p_1$ at slice one.\\
We choose the gamma distribution for $f_{u0}$, $f_{u1}$ and $f_n$ in~\eqref{eq:pdf_jump}, which has support of $(0,\infty)$. For the choice of other distributions the corresponding density functions has to be replaced in \eqref{eq:ev_x0}, \eqref{eq:ev_x1} and the following likelihood formulation \eqref{eq:jump_likelihood_full}, would need to be adapted. The gamma distribution has inverse scale and shape parameters $\beta$ and $k$, respectively. We let $k$ be determined globally and adjust $\beta$ per sample such that the distribution expected value satisfies~\eqref{eq:calc_EX0_EX1} and ~\eqref{eq:Eu0}. Define $k=\mu(p_1)\cdot\beta$, then $\mu(p_1)$ is the desired mean for the corresponding distributions.\\
To summarize: This procedure lets us model the critical outer slices, which are difficult even for human experts, and hence can induce large prediction errors when significant large heart areas are misclassified. The introduced distribution can model those particular errors, while remaining bias free with respect to neural network predictions. The model also remains non-negative and prevents zero volume between non-zero slices, which would lead to disconnected volumes throughout the heart.%
\begin{figure}[tbhp!]
    \begin{center}
        \input{figures/single_sample_with_simulations_vs_1.pgf}

    \end{center}
    \label{fig:single_sample}
    \caption{The results of the proposed method on two different sized hearts. On the left: The neural network prediction $p_t$ in blue. The shaded areas depict the quantiles of 10,000 paths, containing 50\%, 90\% and 99\% of the paths. On the right: A kernel density estimate of the resulting volume prediction of these 10,000 paths. The green shades correspond to the same quantiles as on the left.}
\end{figure}
\subsection{Combining SDE and jump distribution} \label{ssec:combining_sde_j} Combining the SDE for intermediate slices and the jump distribution for apical and basal sections requires these parts to be independent, since errors in the respective areas are presumably caused by independent underlying error sources.\\
The SDE requires prediction $(p_t)_{t=0,\ldots,N}$ to be continuous and differentiable: As noted earlier: we define $p(t)$ as the linear interpolation between points, and we need to keep the integral of $p(t)$ over $t$ the same as the sum over all points. Thus, we need to ensure we start and end with 0, which is achieved by appending $p_{-1}=p_{N+1}=0$. This leads to, 
\begin{equation}\label{eq:interpolate_bias_free}
\begin{split}
        \int_{-1}^{N+1}p(t)dt&= \sum_{i=-1}^N\int_i^{i+1}p(t)dt\\
        &= \sum_{i=-1}^N\left( p_i + \frac{p_{i+1}-p_i}{2}\right)\\
        &=\frac{p_0}{2}+\frac{p_{N+1}}{2}+\sum_{i=0}^Np_i\\
        &=\sum_{i=0}^Np_i.
\end{split}
\end{equation}

\begin{algorithm}[tbhp!]
\caption{Simulation}
\label{alg:inference}
\begin{algorithmic}
\REQUIRE{Preprocessed predictions $P=p_{0},\ldots ,p_{N}$}
\REQUIRE{Parameters $\lambda_u$, $\lambda_d$, $\beta_{u0}$, $\beta_{u1}$, $\beta_{n}$, $\alpha$, $\theta_0$, $\delta$}
\STATE{\textbf{Calculate} $\theta_t$ for $t\in[2,N-2]$}
\STATE{Set $x_{-1}=0, x_{N+1}=0$}
\FOR{desired number of simulations}
\STATE{\textbf{Draw} uniformly from:}
\begin{ALC@g}
\STATE{\textbf{Jump down} with probability $\lambda_d$}
\begin{ALC@g}
\STATE{$x_0=0$, $x_1=0$}
\end{ALC@g}
\STATE{\textbf{Jump up} with probability $\lambda_u$}
\begin{ALC@g}
\STATE{$x_0\sim \mathrm{Gamma}\left(\frac{\lambda_d \cdot \beta_{u0}}{\lambda_u } p_1,\beta_{u0}\right)$, $x_1\sim \mathrm{Gamma}\left( p_1\cdot\beta_{u0},\beta_{u0}\right)$}
\end{ALC@g}
\STATE{\textbf{No jump} with probability $1-\lambda_u-\lambda_d$}
\begin{ALC@g}
\STATE{$x_0=0$, $x_1\sim \mathrm{Gamma}\left( p_1\cdot\beta_{n},\beta_{n}\right)$}
\end{ALC@g}
\end{ALC@g}
\STATE{\textbf{Draw} uniformly from:}
\begin{ALC@g}
\STATE{\textbf{Jump down} with probability $\lambda_d$}
\begin{ALC@g}
\STATE{$x_N=0$, $x_{N-1}=0$}
\end{ALC@g}
\STATE{\textbf{Jump up} with probability $\lambda_u$}
\begin{ALC@g}
\STATE{$x_N\sim \mathrm{Gamma}\left(\frac{\lambda_d\cdot\beta_{u0}}{\lambda_u } p_{N-1},\beta_{u0}\right)$, $x_{N-1}\sim \mathrm{Gamma}\left( p_{N-1}\cdot\beta_{u0},\beta_{u0}\right)$}
\end{ALC@g}
\STATE{\textbf{No jump} with probability $1-\lambda_u-\lambda_d$}
\begin{ALC@g}
\STATE{$x_N=0$, $x_{N-1}\sim \mathrm{Gamma}\left( p_{N-1}\cdot\beta_{n},\beta_{n}\right)$}
\end{ALC@g}
\end{ALC@g}
\STATE{\textbf{Simulate} starting point $\Tilde{x}_2$:}
\begin{ALC@g}
\STATE{Simulate $dX_t = \theta_t(X_t-p_2)dt+\sqrt{2\alpha \theta_0 X_t}dW_t$ for  $t\in[2-\delta,2]$ with $x_{2-\delta}=p_2$}
\STATE{Set $\Tilde{x}_2=x_2$}
\end{ALC@g}
\STATE{\textbf{Simulate} intermediate slices}
\begin{ALC@g}
\STATE{Define $p_t$ in between the points $i$ and $i+1$ as the linear interpolation of $p_i$ and $p_{i+1}$.}
\STATE{Simulate $dX_t = (\dot{p}-\theta_t(X_t-p_t))dt+\sqrt{2\alpha \theta_0 X_t}dW_t$ for  $t\in[2,N-2]$ with $x_2=\Tilde{x}_2$}
\end{ALC@g}
\STATE{$Volume=\int_{-1}^{N+1}x_tdt$}
\ENDFOR
\end{algorithmic}
\end{algorithm}
For each simulation $x(t)$ for $t\in [-1,N+1]$ we first draw values  $x_0, x_1, x_{N-1}, x_N$ from the jump distribution as described in ~\cref{ssec:wall_modelling}. We require a starting value to initialize the SDE, where $p_1$ is not optimal, since the SDE assumes consistent error transitions, but the error at $p_1$ is caused by a different mechanism. Starting at $p_2$ would mean the model always has $x_2=p_2$, which is also undesirable because we also want to model deviations from prediction at slice two. Therefore, we introduce an additional parameter $\delta$, following~\cite[Sec.~4.5]{caballero_quantifying_2021}. To ensure the model has similar error behavior for the second slice as subsequent slices, we start the process at an assumed point $\tilde{p}_{2-\delta}=p_2$ at $t=2-\delta$, and  evaluate the SDE
\begin{equation}\label{eq:SDE_in_procedure}
    \begin{cases}
        dX_t = (\dot{p}(t)-\theta_t(X_t-p_t))dt+\sqrt{2\alpha \theta_0 X_t}dW_t,\\
        t\in [2-\delta,N-2], X_{2-\delta} = p_2 > 0,
    \end{cases}
\end{equation}
where $p_t$ is the linear interpolation of $\tilde{p}_{2-\delta},p_2,p_3\ldots,p_{N-2}$ for $t\in[2-\delta,N-2]$.\\
We integrate $x(t)$ over $t$ to obtain the volume. For $t\in [-1,1]\cup [N-2,N+1]$ $x(t)$ contains the linear interpolation of $x_{-1}=0,x_0,x_1,x_2$ and $x_{N-2},x_{N-1},x_N, x_{N+1}=0$. Between these piece-wise linear functions $x(t)$, $t\in(2,N-2)$, is given by one simulation of the SDE \eqref{eq:SDE_in_procedure}. In Figure~\ref{fig:single_sample} we can see the bands created by 10,000 trajectories of the combined approach with SDE and jump distribution along with the final volume prediction.

\section{Likelihood Formulation}\label{sec:likelihood}
We adopt a maximum likelihood approach to find our model parameters, which requires parameter likelihoods, depending on the segmentation algorithm predictions $P^{(j)}=\{p_{0}^{(j)},\ldots,p_{N_j}^{(j)}\}$ and labeled ground-truth data $G^{(j)}=\{g_{0}^{(j)},\ldots,g_{N_j}^{(j)}\}$, with $p_i^{(j)},g_i^{(j)}\geq 0$ and $j\in \{1,\ldots,M\}$. Algorithm~\ref{alg:fitting} shows the proposed procedure.\\
The complete likelihood $\mathcal{L}_C$ factors nicely into three parts: $\mathcal{L}_{Jump}$ for the first and last two slices, $\mathcal{L}_{SDE}$ for inner slices and $\mathcal{L}_{\delta}$ for the connection between the first jump and the inner slices:
\begin{equation}
	\mathcal{L}_{C}=\mathcal{L}_{Jump}\left(\Theta; G, P\right) \mathcal{L}_{SDE}\left(\Theta; G, P\right)\mathcal{L}_{\delta}\left(\Theta; G, P\right),
\end{equation}
with $\Theta$ being the set of all parameters.

\subsection{Likelihood of the Jump Parameters}
First, consider the jump parameters $\Theta_J=(\lambda_d,\lambda_u,\beta_{u0},\beta_{u1},\beta_n)$. Jump parameter likelihoods on the observed dataset $G,P$ with $M$ observations can be expressed as 
\begin{equation}
    \mathcal{L}_{Jump}\left(\Theta_J; G, P\right) = \prod_{i=1}^{M} \left(f_{X_0,X_1}\left(g_0^{(i)}, g_1^{(i)}\,\middle\vert\ p_{1}^{(i)}, \Theta_J\right)\cdot f_{X_0,X_1}\left(g_N^{(i)}, g_{N-1}^{(i)}\,\middle\vert\ p_{N-1}^{(i)}, \Theta_J\right)\right)
\end{equation}
where $f_{X_0,X_1}$ is the joint jump distribution \eqref{eq:pdf_jump_closed}. Initial and final jumps share the same parameters. Therefore, reindexing the ground-truth labels for the latter edge as $g_0^{(i+M)} \coloneqq g_N^{(i)}$, $g_1^{(i+M)} \coloneqq g_{N-1}^{(i)}$ and $p_1^{(i+M)} \coloneqq p_{N-1}^{(i)}$ can simplify the likelihood notation to \begin{equation}\label{eq:jump_likelihood_first}
    \mathcal{L}_{Jump}\left(\Theta_J; G^{(i)}, P^{(i)}\right) = \prod_{i=1}^{2M} f_{X_0,X_1}\left(g_0^{(i)}, g_1^{(i)}\,\middle\vert\ p_{1}^{(i)}, \Theta_J\right).
\end{equation}
Now consider mutually disjoint index sets $I_n,I_u,I_d\subset \{1,\ldots ,2M\}$ with $I_n \cup I_u \cup I_d = \{1,\ldots ,2M\}$. Each set contains the indices for the different jump cases: no jump, jump up, jump down. Hence for $i\in I_d \Rightarrow g_0^{(i)}=0, g_1^{(i)}>0$, $i\in I_u \Rightarrow g_0^{(i)},g_1^{(i)}>0$ and $i\in I_d \Rightarrow g_0^{(i)}=g_1^{(i)}=0$. There is no observation $i$ in the dataset with $g_0^{(i)}>0$ and $g_1^{(i)}=0$, which confirms our assumption that the LV volume cannot be interrupted with slices with no volume. Therefore, the three cases above cover all possible cases and are mutually exclusive. Thus, using those index-sets and \eqref{eq:pdf_jump_closed}, we can express \eqref{eq:jump_likelihood_first} as
\begin{equation}\label{eq:jump_likelihood_full}
\begin{split}
    \mathcal{L}_{Jump}\left(\Theta_J; G^{(i)}, P^{(i)}\right) =& \prod_{i\in I_d} \lambda_d \cdot  \prod_{i\in I_u} \lambda_u f_{u0}(g_0^{(i)}\mid\beta_{u0})\cdot f_{u1}(g_1^{(i)}\mid\beta_{u1}) \\
    &\cdot  \prod_{i\in I_n}(1-\lambda_d-\lambda_u)  f_n(g_1^{(i)}\mid\beta_n).\\
    =&\lambda_d^{|I_d|} \cdot  \lambda_u^{|I_u|} \cdot (1-\lambda_d-\lambda_u)^{2M-|I_d|-|I_u| }\\ &\cdot \prod_{i\in I_u} f_{u0}(g_0^{(i)}\mid\beta_{u0})\cdot f_{u1}(g_1^{(i)}\mid\beta_{u1})\cdot  \prod_{i\in I_n}  f_n(g_1^{(i)}\mid\beta_n).
\end{split}
\end{equation}
Recall that $f_{u0}$, $f_{u1}$, $f_n$ are density functions of the gamma distributions defined for the different jump cases in \cref{ssec:wall_modelling}. Our goal is to find parameters $\Theta_J$ that maximize likelihood $\mathcal{L}_{Jump}\left(\Theta_J; G^{(i)}, P^{(i)}\right)$. Using the formulation in \eqref{eq:jump_likelihood_full} and knowing that all factors are greater than zero, we can separate the likelihood into two maximization problems. One for $\lambda_u$ and $\lambda_d$ and one for the parameters $\beta_{u0}, \beta_{u1}$ and $\beta_{n}.$
First consider the lambdas,
\begin{equation}\label{eq:lambda_max_likelihood}
    \underset{\substack{\lambda_u\geq0,\lambda_d\geq0\\ \lambda_u +\lambda_d\leq1}}{\mathrm{argmax}}\lambda_d^{|I_d|} \cdot  \lambda_u^{|I_u|} \cdot (1-\lambda_d-\lambda_u)^{2M-|I_d|-|I_u|}.
\end{equation}
The optimal solution is $\lambda_d = \frac{|I_d|}{2M}$ and   $\lambda_u = \frac{|I_u|}{2M}$, as can be seen by maximizing the logarithm of \eqref{eq:lambda_max_likelihood}. It follows that the model's probability for jump ups and downs must be the same as the observed frequency of the respective jumps in the data.\\
The likelihood for $\beta_{u0}, \beta_{u1}$ and $\beta_{n}$ can be determined by the likelihood of the gamma distributions $u_0$, $u_1$, and $n$. Note that the likelihood can also be separated for each of the distributions.\\
\subsection{Likelihood of the SDE Parameters}
For the SDE parameters $\Theta_{SDE}=\{\alpha, \theta_0\}$, we can formulate the likelihood as a product of transition densities in the V-space (see \eqref{eq:SDE_V}). Therefore, we transfer the labeled ground-truth data into V-space $V^{(i)}=(v_0^{(i)},v_1^{(i)},\ldots v_{N_i}^{(i)})$, with $v_j^{(i)}=g_j^{(i)}-p_j^{(i)}$, for MRI  $i\in\{1,\ldots, M\}$. Similar to (12) in~\cite{caballero_quantifying_2021}, we define a transitional density $\rho(v\mid v^{(i)}_{j-1};\Theta_{SDE})$. Then the likelihood function for $V=V^{(0)},\ldots,V^{(M)}$ can be expressed as
\begin{equation}\label{eq:likelihood_in_V}
    \mathcal{L}_{SDE}(\Theta_{SDE};V)=\prod_{i=1}^M\left(\prod_{j=3}^{N_i-2} \rho(v_j^{(i)}\mid v_{j-1}^{(i)};p^{(i)}_{[j-1:j]},\Theta_{SDE}) \right).
\end{equation}
We would need to solve an initial-boundary value problem to derive an exact solution for the transition densities (see (13) in~\cite{caballero_quantifying_2021}), which would be computationally expensive. To avoid this, we approximate the likelihood similar to~\cite[Sec.~4.2]{caballero_quantifying_2021}, selecting the gamma distribution as the surrogate transition density, since this is the invariant distribution of \eqref{eq:SDE_V}. For $t\in [j,j+1)$ we can describe the first two moments $m_1(t)=\mathbb{E}[V_t]$ and $m_2(t)=\mathbb{E}[V_t^2]$ with the initial value problem as follows
\begin{equation}\label{eq:Moment_matching_ode}
    \begin{cases}
        \frac{dm_1(t)}{dt} = -m_1 (t)\theta_t\\
        \frac{dm_2(t)}{dt} = -2\theta_t m_2 (t) + 2\alpha\theta_0 m_1(t)+2\alpha\theta_0 p_t,
    \end{cases}
\end{equation}
with initial conditions $m_1(j)=v_j$ and $m_2(j)=v_j^2$. The system \eqref{eq:Moment_matching_ode} can be deduced using It\^{o}'s lemma. We obtain shape and scale parameters $(k,\beta_{\gamma})$ by matching the gamma distribution with the first two moments,
\begin{equation*}
    \begin{split}
        \beta_{\gamma}(j+1) &= \frac{\mu_{j+1}}{\sigma_{j+1}^2}\\
        k(j+1) &=\frac{\mu_{j+1}^2}{\sigma_{j+1}^2},
    \end{split}
\end{equation*}
where $\mu_{j+1}=m_1(j+1)+p_{j+1}$, $\sigma_{j+1}^2=m_2(j+1)-m_1(j+1)^2$ we shifted the mean by $p_t$, to translate back to the X-space. This is possible as the variance $\sigma_{j+1}^2$ is invariant to changes in location, hence we can now express \eqref{eq:likelihood_in_V} as an approximate likelihood $\tilde{\mathcal{L}}_{SDE}$,
\begin{equation}\label{eq:likelihood_in_X_matched}
    \tilde{\mathcal{L}}_{SDE}(\Theta_{SDE};G)=\prod_{i=1}^M\left(\prod_{j=3}^{N_i-2} f_{\gamma}(g_j^{(i)}\mid \beta_{\gamma}(j^-),k(j^-)) \right),
\end{equation}
where $f_{\gamma}$ is the probability density function, and  shape $k(j^-)$ and scale $\theta(j^-)$ parameters solely depend on the limits $\mu(j^-,\Theta_{SDE})$ and $\sigma(j^-,\Theta_{SDE})$ for $t\uparrow j$, which can computed by numerically solving the initial value problem~\eqref{eq:Moment_matching_ode}.\footnote{The notation $f(a^-)$ describes the one-sided limit $\lim_{x\uparrow a}f(x)$ from below.} This enables computing an approximate likelihood for SDE parameters $\Theta_{SDE}$, and we find the optimal parameters by minimizing the negative logarithm of \eqref{eq:likelihood_in_X_matched} with the gradient free downhill simplex algorithm implemented in SciPy.\\
\begin{algorithm}[tbhp!]
\caption{Fitting}
\label{alg:fitting}
\begin{algorithmic}
\REQUIRE{Preprocessed predictions $P^{(j)}=p_{0}^{(j)},\ldots ,p_{N_j}^{(j)}$, $j=1,\ldots,M$}
\REQUIRE{Preprocessed ground-truth $G^{(j)}=g_{0}^{(j)},\ldots ,g_{N_j}^{(j)}$, $j=1,\ldots,M$}
\STATE{\textbf{Compute} jump parameters $\lambda_u$, $\lambda_d$, $\beta_{u0}$, $\beta_{u1}$, $\beta_{n}$:}
\begin{ALC@g}
\STATE{Find index sets $I_n, I_u, I_d$ as in \eqref{eq:jump_likelihood_full}}
\STATE{$\lambda_n=\frac{|I_n|}{2M}$, $\lambda_u=\frac{|I_u|}{2M}$, $\lambda_d=\frac{|I_d|}{2M}$ according to \eqref{eq:lambda_max_likelihood}}
\STATE{Use gradient-free maximization scheme to find $\beta_{u0}$, $\beta_{u1}$, $\beta_{n}$ with \eqref{eq:jump_likelihood_full}:}
\begin{equation*}
    \underset{\beta_{u0}, \beta_{u1}, \beta_{n}}{\mathrm{argmax}}\prod_{i\in I_u} f_{u0}(g_0^{(i)}\mid\beta_{u0})\cdot f_{u1}(g_1^{(i)}\mid\beta_{u1})\cdot  \prod_{i\in I_n}  f_n(g_1^{(i)}\mid\beta_n)
\end{equation*}
\end{ALC@g}
\STATE{\textbf{Compute} SDE parameters $\alpha$, $\theta_0$:}
\begin{ALC@g}
\STATE{Use gradient free maximization scheme to maximize approximate likelihood \eqref{eq:likelihood_in_X_matched}:}
\begin{equation*}
    \underset{\alpha, \theta_0}{\mathrm{argmax}}\prod_{i=1}^M\left(\prod_{j=3}^{N_i-2} f_{\gamma}(g_j^{(i)}\mid \beta_{\gamma}(j^-),k(j^-)) \right)
\end{equation*}
\STATE{$\beta_{\gamma}(j^-),k(j^-)$ are obtain by solving the initial value problem \eqref{eq:Moment_matching_ode} for each slice $j$}
\end{ALC@g}
\STATE{\textbf{Compute} combining parameter $\delta$:}
\begin{ALC@g}
\STATE{Use gradient free maximization scheme to maximize the approximate likelihood \eqref{eq:approx_likelihood_delta}}
\end{ALC@g}
\end{algorithmic}
\end{algorithm}
Likelihood $\mathcal{L}_\delta$ for $\delta$ can be approximated similar to $\mathcal{L}_{SDE}$. Setting $\tilde{v}_{2-\delta}=0$, as described in \cref{ssec:combining_sde_j}, and again assuming a gamma distribution as transition density $\rho(v_2 \mid v_{2-\delta};p_{2},\Theta_{SDE})$. We approximate this transition density with the same moment matching method as in \eqref{eq:Moment_matching_ode} to obtain the approximate likelihood $\tilde{\mathcal{L}}_{\delta}(\delta;G,\Theta_{SDE})$,
\begin{equation}\label{eq:approx_likelihood_delta}
    \tilde{\mathcal{L}}_{\delta}(\delta;G,\Theta_{SDE})=\prod_{i=1}^Mf_{\gamma}(g_2^{(i)}\mid \beta_{\gamma}(2^-),k(2^-)),
\end{equation}
where $\beta_{\gamma}(2^-),k(2^-)$ are solutions for the initial value problem\eqref{eq:Moment_matching_ode} at $t=2$ starting with $\tilde{v}_{2-\delta}=0$ at $t=2-\delta$. We calibrate $\delta$ after estimating $\Theta_{SDE}$.\\
In contrast to \cite{caballero_quantifying_2021}, our tests were insensitive regarding particular choices of the starting values in the optimization process. However, the method to find initial values from \cite{caballero_quantifying_2021} could be, in principle, adapted to match the SDE \eqref{eq:SDE_V}.
\section{Experimental Results}
\label{sec:exp_results}
The proposed method adds an uncertainty estimate to the point prediction for LV volume, where the point prediction is obtained by an arbitrary machine learning algorithm. The method must capture the uncertainty of the data and adjust for different data regimes. Furthermore, it needs to be stable on smaller datasets.\\
For this evaluation, we used two cardiac MRI datasets: The Multi-Centre, Multi-Vendor, and Multi-Disease Cardiac Segmentation (M\&Ms) dataset~\cite{MnM_2021} and the Automatic Cardiac Diagnosis Challenge (ACDC) dataset~\cite{Acdc_2018}. The ACDC dataset comprises 300 MRIs from 150 patients, where each patient has one MRI in the end-diastolic and one in the end-systolic phase. Patients are evenly distributed between five classes: four pathologies and one extra for healthy patients. All images come from the same research center. In contrast to the ACDC dataset, the M\&Ms dataset focuses on generalizing across different medical centers with different scanner vendors. The 375 MRIs comprise images from 6 medical centers with five distinct MRI devices. Pathologies are considerably less evenly distributed in this dataset. Although healthy patients comprise one-third of the data, the second largest pathology class (hypertrophic cardiomyopathy) also contains 103 MRIs. Thus, more than 60\% of the data are in the two largest groups. We only used the training sets from both datasets because labels are only available for the training sets.\footnote{Final ACDC dataset contains 200 MRIs from 100 patients, and M\&Ms dataset contains 300 MRIs from 150 patients.}\\%
\begin{figure}[tbhp!]%
    \begin{center}%
        \input{figures/error_full_heart.pgf}%
    \end{center}%
    \label{fig:error_full_heart}%
    \caption{Comparison of the different model-based and empirical cumulative density functions of the deviation from the neural network prediction. Green shows the deviation of the expert labels from the neural network predictions. Light green shows the KS confidence band, based on the DKW-inequality, which includes the true cdf with a probability of 95\%. Orange shows the cdf of our model and blue the cdf of the benchmark model, utilizing the five prediction of the neural network ensemble. To obtain the cdfs the LV volume of each heart is predicted 10,000 times. The benchmark model utilizes the predictions of all five neural networks from the nnU-Net. }%
\end{figure}%
We selected the nnUNet~\cite{isensee_nnu-net_2021} as the pre-trained predictor.\footnote{See \href{https://github.com/MIC-DKFZ/nnUNet}{https://github.com/MIC-DKFZ/nnUNet} for implementation details.} The nnUNet method automatically finds suitable U-Net structures for a wide range of medical datasets and won both ACDC segmentation and M\&Ms challenges. We used the version trained on the more diverse M\&Ms dataset and evaluated the method on the ACDC dataset, providing a whole MRI image dataset unseen during training. The proposed procedure is applicable for real-world applications, where one would select a suitable pre-trained net on the most diverse dataset possible and subsequently evaluate on data from one medical center. Although nnUNet uses a five neural network ensemble for final prediction, we used only the first of those neural networks for simplicity. Real-world use cases would improve prediction quality by using the full ensemble and the largest dataset possible, e.g. combining publicly available M\&Ms and ACDC datasets.%
\subsection{Model Assessment}
As described earlier the nnU-Net employs an ensemble of five neural networks trained on different folds of the training data. Leveraging the predictions of different networks for uncertainty estimation, could be an alternative approach for estimating the uncertainty. As a benchmark model, we will use the five predictions of the full nnU-Net. We will view the prediction of each neural network as a point prediction as we saw in Figure~\ref{fig:error_vs_std} that the variance within each prediction is negligible. We obtain five point predictions for the volume of each slice. We then assume that the volume of each slice is normally distributed with the empirical mean and standard deviation of those five predictions. Let $N+1$ be the number of slices for one heart and $\mu_i,\sigma_i$ be the mean and standard deviation of the five predictions at slice $i\in \{0,\ldots,N\}$. The model for the volume of the full LV is a sum of normal distributions. Hence, it is normally distributed with mean $\sum_{i=0}^N \mu_i$ and variance $\sum_{i=0}^N \sigma_i^2$.%
\begin{figure}[tbhp!]%
    \begin{center}%
        \input{figures/jump_simulation_vs_gt.pgf}%
    \end{center}%
    \label{fig:jump_simulation_vs_gt}%
    \caption{Comparison of the volumes of the edge slice. In order to make the volumes comparable it is normalized by the predicted volume. Blue shows the ground-truth volume and red shows the simulation volumes. For the simulation each heart is evaluated 1000 times. In both the data and the simulation one can see a multimodal distribution in the particularly difficult edge slices.}%
\end{figure}\phantom{-}\\
For our model the parameters were fitted on the whole training dataset, and Section~\ref{ssec:stab_analysis} shows that fitted parameters do not substantially differ on data subsets, confirming that models fitted on data subsets  exhibit similar characteristics. We sample 10,000 volume predictions based on our uncertainty model and the benchmark model for each heart and compared them to labeled ground-truth data. We compare the deviation of the models from the predictions of the neural network with the deviation of the expert labels from the predictions of the neural network. To compare this we corrected the bias of the neural network as we do not want to capture this uncertainty in our model. The bias in the model would need to be addressed separately and is beyond the scope of this work. We use the empirical distribution functions for the comparison to be independent of the choice of bins. To account for the small sample size of the labeled data we also plotted the Kolmogorov--Smirnov (KS) confidence band for the level $\lambda = 0.05$ around the empirical cdf. The KS  confidence band utilizes the Dvoretzky--Kiefer--Wolfowitz (DKW) inequality to estimate the interval around the empirical cdf, that contains the true cdf with probability $1-\lambda$ \cite{massart_tight_1990, dkw_inequality_1956}.\footnote{Let $F(x)$ be the true cdf and $\hat{F}_n(x)$ the empirical cdf for $n$ samples. From the DKW inequality follows with probability $1-\lambda$, that for every $x$, the interval that contains the true cdf $F(x)$ is a subset of $[\hat{F}_n(x)-c_{n,\lambda},\hat{F}_n(x)+c_{n,\lambda}]$, where $c_{n,\lambda}=\sqrt{ln(2/\lambda)/(2n)}$.} In Figure~\ref{fig:error_full_heart} we do not only see, that our model is a lot better in describing the uncertainty seen in the data. It also is able to capture the data quite well including the width of the deviation from the neural network prediction. Figure~\ref{fig:error_full_heart} shows that our method captures the error of the full heart well. It lays well within the KS confidence band. To investigate further, we examined the volume of the difficult edge slices. Because of the different volumes, it is hard to compare the data over several hearts. To circumvent this problem we divided the expert labeled volume and the predicted volume of our model, by the predicted volume of the neural network. This way for each heart the expected volume is one and we can see the multimodality of the jumps in the data. Figure~\ref{fig:jump_simulation_vs_gt} shows, that our approach of modelling the jumps captures this multimodality well. This multimodality is mostly evident for total volume predictions of one heart where the edge slices have significant influence on overall volume.%
\begin{figure}[tbhp!]%
    \begin{center}%
        \input{figures/at_vs_volume.pgf}%
    \end{center}%
    \label{fig:at_vs_volume}%
    \caption{Heart size effects on $\alpha \theta_0$ obtained by maximizing the likelihood on a subset of the data. The subsets are binned by volume, with each bin receiving approximately the same number of observations: (left) 8 bins, (right) 16 bins. The scatter points are located at the midpoint for the respective bin. This shows that for larger volumes the optimal $\alpha\theta_0$ get smaller. This trend is consistent, even when very small subsets of data are used.}%
\end{figure}\phantom{-}\\%
Additional to the difficult edge slices the ACDC dataset contains significant heart variance due to different pathologies and end-systolic and end-diastolic phases included in the data. The LV volumes range from $\sim\!25$ to $\sim\!325$~ml. Therefore, we investigated model effects due to different heart sizes. The hearts were binned by predicted volume, so we could also use this binning for the inference step, where no ground-truth label is available. We then calculated parameters $\alpha_i$ and $\theta_{0,i}$ for each bin $i$. Figure~\ref{fig:at_vs_volume} confirms $\alpha_i \theta_{0,i}$ decrease consistently with LV volume. The trend remains consistent up to 16 bins, providing only 12 or 13 observations per bin. Larger values for $\alpha_i \theta_{0,i}$ in small hearts lead to a higher path variability and uncertainty estimation for those hearts. These outcomes agree well with the larger relative errors in this data regime, see Figure~\ref{fig:small_vs_large}. For smaller hearts the model fitted on the whole dataset underestimates the errors, while still being in the 
KS confidence band for the level $\lambda = 0.05$. The model fitted on the subsets of the small hearts is closer to the empirical cdf in this regime. One could potentially compensate for pre-trained net shortcomings, i.e., higher uncertainty for smaller hearts in this case, by adjusting the predicted uncertainty accordingly.%
\begin{figure}[tbhp!]%
    \begin{center}%
        \input{figures/small_vs_large.pgf}%
    \end{center}%
    \label{fig:small_vs_large}%
    \caption{Comparison of the different model-based and empirical cumulative density functions of the deviation from the neural network prediction on the 25 smallest hearts. Green shows the deviation of the expert labels from the neural network predictions. Light green shows the KS confidence band, based on the DKW-inequality, which includes the true cdf with a probability of 95\%. Orange shows the cdf of the model fitted on the full dataset and blue the model fitted on only the small models. While both models lie within the KS confidence band, which is quite wide as the dataset only consists of 25 samples, the model fitted on the small hearts is better at capturing the relative larger deviation for small hearts.}%
\end{figure}%
\FloatBarrier
\subsection{Stability analysis}\label{ssec:stab_analysis}
We evaluated the proposed method's stability by assessing parameter stability to data changes. Figure~\ref{fig:nll_sde_params} shows the negative log-likelihood for SDE parameters $\alpha$ and $\theta_0$ with respect to the entire dataset. The minimum for the relevant parameter range is quite wide, and there are no other local minima.\begin{figure}[tbhp!]
    \begin{center}
        \input{figures/nll_sde_parameters.pgf}
    \end{center}
    \label{fig:nll_sde_params}
    \caption{Fitted parameters for different data subsets. The full dataset was divided three times into three parts, obtaining nine datasets, with always three being disjoint. The same procedure was performed on the dataset with reversed order for the slices for each heart. The contour plot shows negative log-likelihood with respect to the full standard dataset.}
\end{figure} To further study the proposed procedure stability on smaller datasets, we divided the dataset into three disjoint subsets with 64 MRIs in each, then repeated this procedure three times to obtain nine subsets, with always three being pairwise disjoint. Furthermore, our method allows to choose the direction we go through the MRI slices, since we do not have a preferred or optimal direction, i.e., from basal to apical sections or reverse. Therefore, we also generated nine subsets for the reversed dataset following the same procedure. We see that the parameters for all subsets lie in the same region with similar likelihood. This implies that the procedure is stable on the subsets of data. The consistent trend when using small bins of 12--13 observations in Figure~\ref{fig:at_vs_volume} implies stability for even smaller datasets.\\
The main challenge for assessing optimal jump parameters is that jumps are scarce in the data. A jump down occurred for approximately $5.4\%$ of the outer slices, with the jump-up case slightly more frequent (approximately $7\%$). Therefore, we modeled jumps in basal and apical sections with the same parameters, $\lambda_u$ and $\lambda_d$, i.e., observed jump frequency in the dataset. Assuming true jump frequency of $6\%$, we calculated the observed frequency for a given subset with the binomial distribution as shown in Figure~\ref{fig:lambda_on_subsets}.%
\begin{figure}[tbhp!]%
    \begin{center}%
        \input{figures/lambda_on_subsets.pgf}%
    \end{center}%
    \label{fig:lambda_on_subsets}%
    \caption{Probability of estimating different $\lambda$ for different sample sizes, assuming true $\lambda_{\rm true}=0.06$.}%
\end{figure}To investigate parameter stability for describing gamma distributions, we again split the full dataset into three parts (64 MRIs) and found maximum likelihood parameters for each subset. Figure~\ref{fig:gamma_theta_on_subsets} shows outcomes after repeating this process ten times. Although jumps remain quite rare, parameters for the jump distributions are stable across the various data subsets, with only one potential outlier from 30 subsets.\\
To test for the combined effect of the jump parameters and the SDE paremeters on the stability, we again divided the dataset into 18 subsets, with always three beeing pairwise disjunct and nine of them being reversed. Figure~\ref{fig:full_model_stability} shows, that the models fitted on each of the 18 subsets lay well within the KS confidence band of the empirical CDF of the full dataset. This indicates that the full model, comprised of the jump distribution and the SDE, is stable on small datasets.
\begin{figure}[tbhp!]
    \begin{center}
        \input{figures/subsamples.pgf}
    \end{center}
    \label{fig:full_model_stability}
   \caption{Comparison of the different model-based and empirical cumulative density functions of the deviation from the neural network prediction on the full dataset. Green shows the deviation of the expert labels from the neural network predictions. Light green shows the KS confidence band, based on the DKW-inequality, which includes the true cdf with a probability of 95\%. Orange shows the cdf of the different models fitted on subsets of the data. In total there are 18 subsets of the data with always three being pairwise disjoint and in nine the direction is flipped, cf. Figure~\ref{fig:nll_sde_params}. The models fitted on all subsets stay within the KS confidence band.}
\end{figure}
\section{Conclusion}\label{sec:conclusion}
We developed a phenomenological model to describe the uncertainty for left-ventricle volume prediction. Accurately measuring LV volume is crucial to properly assess cardiovascular mortality or diagnose heart failure. The proposed method is a post-hoc approach to quantify uncertainty for predictions of any segmentation algorithm.\\
Our requirements for the model was that it should
\begin{enumerate}
    \item consists of few parameters, so that it can be fitted on limited data,
    \item stay non-negative and bias-free with respect to the neural network prediction,
    \item and most importantly describe the uncertainty well.
\end{enumerate}
Those requirements should be met with few assumptions. The SDE approach is well suited to model the transitions and naturally model the correlations. We only require two parameters for the SDE to obtain a flexible model, which meets our requirements and is thus very parameter-efficient.\\
Our model employs the inherent locality, to obtain reliable uncertainty estimations, using heart depth as \emph{fake time} to model deviation from underlying deterministic predictions with an SDE. Outer slice errors arise from different sources than inner slices and must be modeled independently. Therefore, we designed a jump distribution to model systematic error in these slices. Finally, we introduced an additional parameter $\delta$ to seamlessly combine the SDE and jump distribution while retaining their independence.\\
Max likelihood approaches were employed to fit the parameters. We described the likelihood using an analytical expression for the jump distribution and constructed an approximate likelihood formulation for the SDE parameters of the model. Moment matching was employed to describe a gamma distribution for the transition density between slices.\\
The proposed method is well suited to describe uncertainty for LV volume prediction, with estimated uncertainty matching ground-truth data errors well. The proposed distribution was able to reproduce the observed multimodality in ground-truth data even for particularly difficult outer slices.\\
There is a wide range of possible phenomenological models. Our approaches meets this applications requirements. Different conceivable approaches include Gaussian processes. Similar to our approach, Gaussian processes typically have few parameters but would need to be modified to fulfill the above mentioned requirements. Especially keeping it non-negative while being bias-free with respect to the neural network prediction is not trivial. A different possible approach is leveraging the flexibility of diffusion models. Diffusion models share some similarities with the SDE approach and show promising results in a wide range of applications. In \cite{tamir_transport_2023} the authors integrate Bayesian filtering into learning the diffusion models to generate stochastic processes governed by sparse observations. The sparse observations could be the predictions of the slices for our case. The time dynamics are also modelled by an Itô SDE. The drift function is typically modelled by a neural network. This increases the flexibility of the model compared to the drift function defined in \eqref{eq:drift}. On the other hand it also drastically increases the number of parameters which need to be fitted, which could be problematic in our use-case. Furthermore, the drift and diffusion function would need to be adapted, such that they ensure non-negativity. The comparison of other phenomenological models such as Gaussian processes and diffusion models with our approach would be an interesting direction of future research.\\
In real-world application, most medical centers only have small labeled datasets and can only use publicly available data with various limitations due to differing vendors and imaging protocols from center to center. The stability of the proposed method with respect to dataset size and its ability to be combined with any segmentation algorithm is crucial and makes the method applicable for real-world scenarios. Our evaluations showed the proposed method is stable even on small datasets.\\
This proposed method can also compare segmentation algorithms performance on different data sub-regimes. We found increased relative error for smaller hearts, which suggests one could adapt this method to account for the different behavior in different data regimes using parameters dependent on the predicted volume.\\
Automatic LV volume prediction would facilitate and speed up diagnosis, but reliable uncertainty quantification is essential. In contrast to current best practice approaches, the proposed adaptable, post-hoc method provides this critical parameter, and paves the way toward reliable automatic left ventricle volume prediction.
\begin{figure}[tbhp!]
    \begin{center}
        \input{figures/gamma_theta_on_subsets.pgf}
    \end{center}
    \label{fig:gamma_theta_on_subsets}
    \caption{Fitted parameters for different data subsets. The full dataset is divided ten times into three parts, providing 30 datasets, with always three being disjoint. Values are shown with the likelihood calculated from the full dataset.}
    \end{figure}
\section*{Acknowledgments}
This publication is partly supported by the Alexander von Humboldt Foundation and the King Abdullah University of Science and Technology (KAUST) Office of Sponsored Research (OSR) under Award No. OSR-2019-CRG8-4033. This work was performed  as  part  of  the  Helmholtz School for Data Science in Life, Earth and Energy (HDS-LEE) and received funding from the Helmholtz Association of German Research Centres.

\bibliographystyle{siamplain}
\bibliography{literature}

\end{document}